\pdfoutput=1
\documentclass[11pt]{article}

\usepackage{EACL2023}

\usepackage{times}
\usepackage{latexsym}

\usepackage[T1]{fontenc}
\usepackage[utf8]{inputenc}

\usepackage{microtype}

\usepackage{inconsolata}

\usepackage{amsmath}
\usepackage{booktabs}
\usepackage{threeparttable}
\usepackage{multirow}
\usepackage[inline]{enumitem}
\usepackage{subcaption}
\usepackage{microtype}
\usepackage{amsthm}
\usepackage{enumitem}
\usepackage[normalem]{ulem}  % Necessary for strikeout, normalem to keep \emph as italics
\usepackage{tikz}
\usepackage{amsfonts}
\usepackage{algorithm}
\usepackage[noend]{algpseudocode}
\usetikzlibrary{bayesnet}
\usepackage{cleveref}
\usepackage{xcolor}
\usepackage[utf8]{inputenc}
\usepackage{arydshln}
\newcommand{\bb}[1]{\mathbf{#1}}

\title{Assistive Recipe Editing through Critiquing}

\author{Diego Antognini\thanks{\ \ Work done while at EPFL and a research stay at~UCSD.}\\
  IBM Research\\
  \texttt{diego.antognini@ibm.com} \\\And
  Shuyang Li\thanks{\ \ Work done while at UCSD.}\\
  Meta AI \\
  \texttt{lishuyang@meta.com} \\\AND
  Boi Faltings\\
  EPFL\\
  \texttt{boi.faltings@epfl.ch} \\\And
  Julian McAuley\\
  UCSD\\
  \texttt{jmcauley@ucsd.edu} \\
}

\begin{document}
\maketitle

\begin{abstract}

Home cooks often have specific requirements regarding individual ingredients in a recipe (e.g.,~allergies).
Substituting ingredients in a recipe can necessitate complex changes to instructions
(e.g.,~replacing chicken with tofu in a stir fry requires pressing the tofu, marinating it for less time, and par-cooking)---which
has thus far hampered efforts to automatically create satisfactory versions of recipes.
We address these challenges with the \emph{RecipeCrit} model that allows users to \emph{edit} existing recipes by proposing individual ingredients to add or remove.
Crucially, we develop an unsupervised critiquing module that allows our model to iteratively re-write recipe instructions to accommodate the complex changes needed for ingredient substitutions.
Experiments on the Recipe1M dataset show that our model can more~effectively edit recipes compared to strong language-modeling baselines, creating recipes that satisfy user constraints and humans deem more~correct, serendipitous, coherent, and relevant.

\end{abstract}

\section{Introduction}

Individual preferences and dietary needs shape the types of recipes that home cooks choose to follow.
Cooks must often accommodate the desire for versions of recipes that do not contain a specific ingredient (substitution---e.g.,~for food allergies) or \emph{do} make use of particular ingredients (addition---e.g.,~to use up near-expiry items).
We thus aim to build a system for \emph{recipe editing} that accommodates fine-grained ingredient preferences.

Prior research in recipe editing has focused on substituting individual ingredients in the ingredients list
\cite{ingr_sub_rec} or recommending new recipes based on
similar ingredients \cite{recipe_rec}.
Individual ingredient substitution rules (e.g.,~tapioca flour and xanthan gum for wheat flour)
often necessitate additional changes to the cooking procedure
to function properly \cite{share}.
Other studies employed recommendation-based approaches.
However, they suffer from data sparsity: there is an extremely large set of possible recipes that differ by a single ingredient, and many specific substitutions may not appear in recipe aggregators \cite{multi_step_demo}.

Recipe editing can be seen as a combination of recipe generation and controllable natural language generation \cite{autoprompt}, and has been explored to adapt recipes for broad dietary constraints \cite{share} and cuisines \cite{cuisine_recipe}.
Pre-trained language models have been used to create recipe directions given a known title and set of ingredients \cite{checklist,bosselut_recipe},
but generated recipes suffer from inconsistencies.
\citet{share} instead build a paired recipe dataset,
but face challenges scaling
due to the large set of possible recipes and dietary restrictions;
people often express even more specific ingredient-level preferences (e.g., dislikes of certain ingredients or allergies). 

In this work, we address the above challenges and propose \textbf{RecipeCrit}, a denoising-based model trained to complete recipes and learn semantic relationships between ingredients and instructions.
The novelty of this work relies on an \emph{unsupervised} critiquing method that allows users to provide ingredient-focused feedback iteratively; the model substitutes ingredients and also re-writes the recipe text using a generative language model.
While existing methods for controllable generation require paired data with
specially constructed prompts \cite{ctrl} or
hyperparameter-sensitive training of 
individual models for each possible piece of feedback \cite{pplm}, our unsupervised critiquing framework enables recipe editing models to be trained with arbitrary un-paired data.
This generalizes recipe editing, unlike existing methods for controllable generation that rely on paired data with specially constructed prompts \cite{ctrl} or hyperparameter-sensitive training of individual models for each possible feedback~\cite{pplm}.

Experiments on the Recipe1M \cite{salvador2017learning} dataset show that RecipeCrit edits recipes in a way that better satisfies user constraints, preserves the original recipe, and produces coherent recipes (i.e.,~recipe instructions are better conditioned on the ingredients list) 
compared to state-of-the-art pre-trained recipe generators and language models.
Human evaluators judge RecipeCrit's recipes to be more serendipitous, correct, coherent, and relevant to the ingredient-specific positive and negative feedback (i.e., critiques).

\section{RecipeCrit: a Hierarchical Denoising Recipe Auto-encoder}
%\section{RecipeCrit Model}
\label{sec_critiquing}

\begin{figure}[t!]
  \centering
  \includegraphics[width=1.0\linewidth]{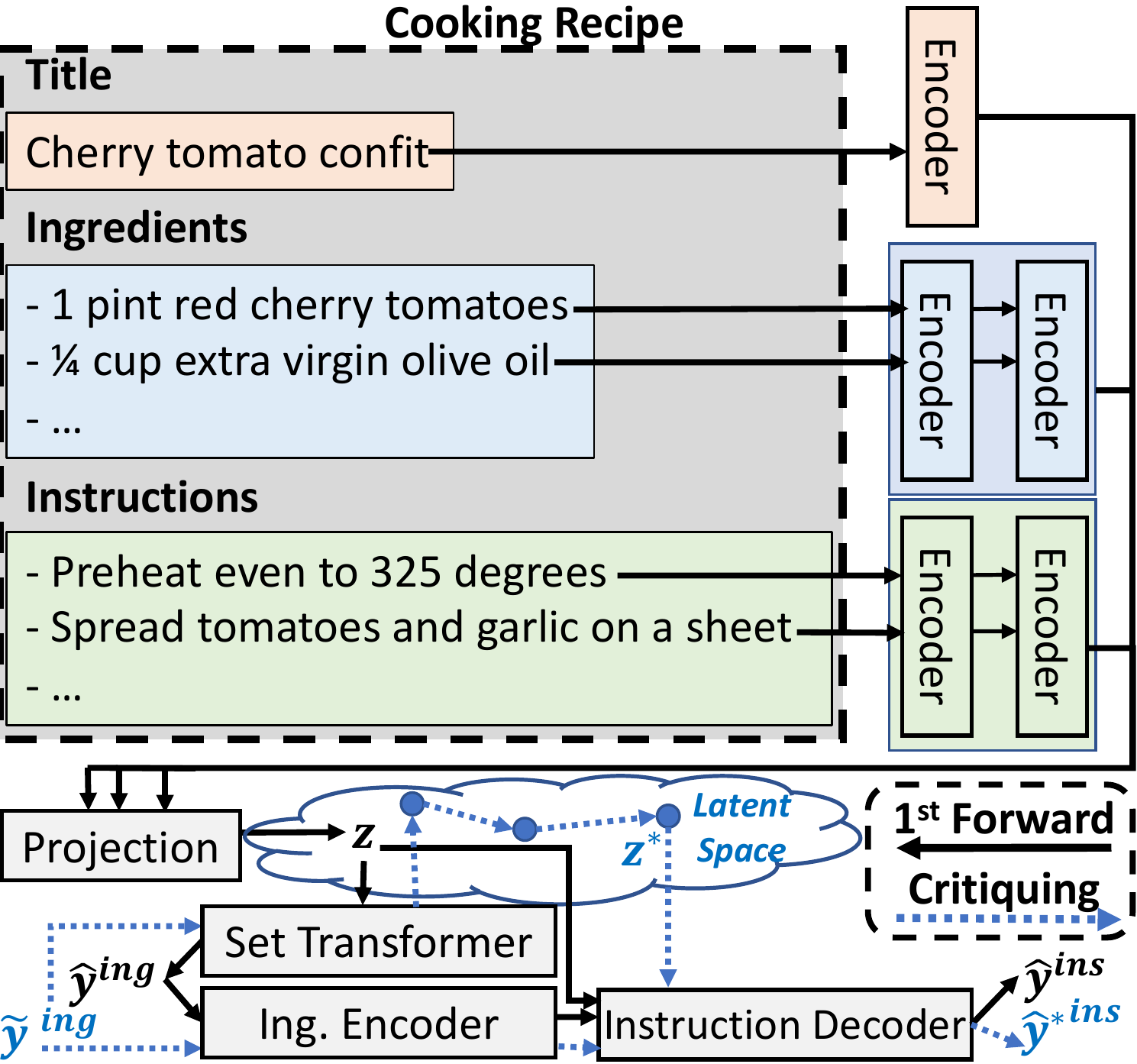}
\caption{RecipeCrit includes a recipe encoder and an ingredient and instruction decoders using the base recipe and target ingredients to edit the instructions.}
\label{fig:model_diagram}
\end{figure}

Previous methods to edit recipes focused on broad classes like dietary categories \cite{share} and cuisines \cite{cuisine_recipe} and require paired corpora (which do not exist for fine-grained edits).
We propose \textbf{RecipeCrit}: a hierarchical denoising recipe auto-encoder that does not require paired corpora to train and accommodates positive and negative user feedback about ingredients (\Cref{fig:model_diagram}).

RecipeCrit is divided into three submodels: an Encoder~$E(\cdot)$, which produces the latent representation $\bb{z}$ from the~(potentially noisy) recipe; an ingredient predictor $C(\cdot)$, which predicts the ingredients $\bb{\hat{y}}^{\mathit{ing}}$, and a decoder $D(\cdot)$, which reconstructs~the cooking instructions $\bb{\hat{y}}^{\mathit{ins}}$ from $\bb{z}$ conditioned on the $\bb{\hat{y}}^{\mathit{ing}}$.

\paragraph{Recipe Encoder $E(\cdot)$}
We build a powerful latent representation that captures the different elements of a recipe via
the mean-pooled output of 
the representation of each sentence using 
a Transformer encoder~\cite{vaswani2017attention}.
We provide the title $\bb{x}^{\mathit{ttl}}$, ingredients $\bb{X}^{\mathit{ing}}$, and instructions $\bb{X}^{inst}$ as \emph{raw} text input.
While the title $\bb{x}^{\mathit{ttl}}$ comprises a single sentence, the ingredients $\bb{X}^{\mathit{ing}}$ and instructions $\bb{X}^{inst}$ are provided as lists of sentences; we use raw recipe texts directly as input, removing the need of a pre-processing step.
We encode the ingredients and instructions in a hierarchical manner using another Transformer to create fixed-length representations.
We compute~the latent representation $\bb{z}$ by concatenating the representations of each component and applying a projection followed by a tanh~function:
$\bb{z} = \textrm{tanh}(\bb{W}[\textrm{TRF}(\bb{x}^{\mathit{ttl}})\mathbin\Vert \textrm{HTRF}(\bb{X}^{\mathit{ing}})\mathbin\Vert \textrm{HTRF}(\bb{X}^{\mathit{ins}})]),$
where $\mathbin\Vert$ is the concatenation, and $\bb{W}$, $\bb{b}$ the projection parameters.

\paragraph{Ingredient Predictor $C(\cdot)$}

We treat ingredient prediction as multi-label binary classification over an ingredient vocabulary $I$, with a boolean target vector $\bb{y}^{\mathit{ing}}$ representing ingredients in a ground truth recipe.
We use a Set Transformer \cite{inverse_cooking} to decode ingredients, pooling ingredient logits over time-steps to compute binary cross-entropy loss against the target; we also employ an \emph{EOS} token to predict the ingredient set cardinality:
 \begin{align*}
 	\mathcal{L}_{\mathit{ing}}(\cdot) = - \sum\nolimits_i^{|I|}\bb{y}^{\mathit{ing}}_i\log \bb{\hat{y}}^{\mathit{ing}}_i
 	- \lambda \bb{y}^{\mathit{ing}}_{eos}\log \bb{\hat{y}}^{\mathit{ing}}_{eos},
 \end{align*}
 where $\lambda$ controls the impact of the \emph{EOS} loss.
At inference, we return the top-k ingredients, where k is the first position with a positive \emph{EOS} prediction.

\paragraph{Instruction Decoder $D(\cdot)$}

The last component generates cooking instructions using a Transformer decoder.
We condition the decoder on $\bb{z}$, previously generated outputs $\bb{\hat{y}}^{\mathit{ins}}_{1:t-1}$, and ingredients $\bb{\hat{y}}^{\mathit{ing}}$.
Specifically, we encode the ingredients using an embedding layer $A(\cdot)$ and concatenate their representations with the recipe representation $\bb{z}$.
We train using teacher-forcing and cross-entropy:%~$\mathcal{L}_{\mathit{ins}}$:
 \begin{equation*}
 		\mathcal{L}_{\mathit{ins}}(\bb{z},A(\bb{\hat{y}}^{\mathit{ing}})) = - \sum\nolimits_t\bb{y}^{\mathit{ins}}_t\log \bb{\hat{y}}^{\mathit{ins}}_t.
\end{equation*}

Taking inspiration from masked language and span modeling \cite{bert,spanbert},
we train RecipeCrit as a de-noising recipe auto-encoder via the task of \emph{recipe completion}:
We mask random ingredients and instruction sentences in model input, and task our model to generate the full recipe.
We train our model in two stages: first minimizing ingredient prediction loss $\mathcal{L}_{\mathit{ing}}$; then freezing the encoder and optimizing for instruction decoding loss $\mathcal{L}_{\mathit{ins}}$ using ground truth ingredients.

\paragraph{Unsupervised Critiquing}

We aim to refine a recipe based on the user's feedback and the predicted ingredients $\bb{\hat{y}}^{\mathit{ing}}$.
We denote $\bb{\tilde{y}}^{\mathit{ing}}$ the vector of desired ingredients. 
Simply incorporating user feedback by explicitly including/removing ingredients before generating instructions often cannot satisfy user preferences due to weak conditioning between predicted ingredients and generated~instructions. %(\Cref{table_crit_perf}).
In RecipeCrit we turn to a~critiquing method that modifies the recipe representation $\bb{z}$ before using the updated representation to jointly generate the edited ingredients and instructions.
Specifically,~users add a new ingredient~$c$ by setting $\bb{\hat{y}}^{\mathit{ing}}_c=1$ or remove some using $\bb{\hat{y}}^{\mathit{ing}}_c=0$.

Inspired by success in~editing the latent space in text style~transfer and recommendation \cite{antognini2020interacting,abs-1905-12926}, we
first compute the gradient with respect to $\bb{z}$: 
 \begin{equation*}
 \bb{g}_{t-1} = \nabla_{\bb{z}_{t-1}} \mathcal{L}_{\mathit{ing}}(C(\bb{z}_{t-1}), \bb{\tilde{y}}^{\mathit{ing}}).
  \end{equation*}
Then ,we use the gradient to modulate $\bb{z}$ such that the new predicted ingredients $\bb{\hat{y}}^{\mathit{ing}}$ are close to the desired ingredients $\bb{\tilde{y}}^{\mathit{ing}}$: 
 \begin{equation*}
 \bb{z}_t = \bb{z}_{t-1} - \alpha_{t-1} \bb{g}_{t-1}/||\bb{g}_{t-1}||_2.
   \end{equation*}
  
Prior work stopped updating $\bb{z}$ when $\parallel\bb{\tilde{y}}^{\mathit{ing}}-\bb{\hat{y}}^{\mathit{ing}}\parallel_1 < \epsilon$ for some threshold $\epsilon$.
We instead propose to compute the absolute difference $|\bb{\tilde{y}}^{\mathit{ing}}_{c}$ - $\bb{\hat{y}}^{\mathit{ing}}_{c}|$.
Since the optimization is nonconvex, we improve convergence by using an early stopping mechanism.
Our approach is unsupervised and~can update the full recipe latent representation, reflecting how adding or removing an ingredient can necessitate adjustments to other ingredients and cooking steps.
Pseudo-code is available in the App. 

Another advantage of our approach is the possibility to update multiple ingredients simultaneously: adding or removing an ingredient might affect other ones as well and thus, a local-based~stopping criteria allows such a change.

\section{Experiments}

\paragraph{Dataset}
\label{sec_datasets}

We assess our model on the Recipe1M \cite{salvador2017learning} dataset
of 1M recipe texts.
Each recipe contains a title, a list of ingredients, and a list of cooking instructions.
We filter out recipes with more than $20$ ingredients or steps, creating train, val, and test splits with 635K, 136K, and 136K recipes, respectively.
The average recipe comprises 9 ingredients and 166 words.
We follow~\citet{inverse_cooking} and build a set of $1,488$ ingredients. For critiquing, we select $20$ ingredients to be critiqued among the most and the least popular~ingredients across the train set.
For each critique, we randomly sample $50$ recipes that contain the critiqued ingredient and $50$ that do not.

\paragraph{Baselines}
\label{sec_baselines}

We compare our proposed RecipeCrit architecture against large language models trained using our denoising objective.
We fine-tune BART \cite{bart}, an encoder-decoder language model trained to denoise documents, as well as RecipeGPT \cite{recipegpt}, a decoder-only language model pre-trained on Recipe1M to predict ingredients and cooking steps.
To demonstrate the necessity of our denoising approach, we also compare against PPLM \cite{pplm}, a recent method for controllable generation from language models that leverages sets of desired and undesired sequences (for ingredient addition and substitution, respectively).
All models use greedy decoding.

\paragraph{Metrics}
\label{sec_metrics}

We evaluate edited recipes via metrics that reflect user preferences.
First, a user wants a recipe similar to the base recipe---we measure \emph{ingredient fidelity} via IoU (Jaccard distance) and F1 scores between the edited and base recipe ingredients list.
Next, the recipe must satisfy the user's specific ingredient feedback---we report the percentage of edited recipes that properly include/exclude the target ingredient (\emph{Success Rate}).
Finally, the recipe must be \emph{coherent}: able to be followed and internally consistent.
As an ingredient constraint can be satisfied in many ways, %we avoid $n$-gram metrics like BLEU \cite{papineni2002bleu}
we follow~\citet{checklist} and measure~coherence via precision, recall, and F1-score of ingredients~mentioned in the generated steps compared to the predicted ingredients.
This verifies that
the recipe itself relies on the listed~ingredients.

\paragraph{Training Details}

For fair comparison, we compare similar-sized models.
RecipeCrit uses an encoder and decoder with 4 Transformer layers, 4~attention heads, and hidden size of 512.
We randomly mask $50\%$ of the ingredients and instructions during training, and tune them on the validation set using random search. We give more details in~App.~\ref{app_training}.

\begin{table}[!t]
\small
    \centering
\begin{tabular}
{@{}
c@{\hspace{1mm}}l@{\hspace{1mm}}
c@{\hspace{1mm}}c@{\hspace{1mm}}c@{\hspace{1mm}}
c@{\hspace{1mm}}
c@{\hspace{1mm}}c@{\hspace{1mm}}c@{}}
& & & \multicolumn{2}{c}{\emph{Ingr. Fidelity}} & & \multicolumn{3}{c}{\emph{Predicted Instr.}}\\
\cmidrule(lr{1em}){4-5}\cmidrule(lr{1em}){7-9}
& \textbf{Model} & \textbf{\% Succ.} & \textbf{IoU} & \textbf{F1} & & \textbf{Prec.} & \textbf{Rec.} & \textbf{F1}\\
\toprule
\multirow{4}{*}{\rotatebox{90}{\emph{Add}}} & RecipeGPT & $33.2$	& $65.4$	& $78.7$	&	& $56.7$	& $69.0$	& $62.2$\\
& PPLM & $34.4$ & $60.9$ & $72.7$ & & $53.0$ & $63.0$ & $57.6$\\
& BART & $41.1$	& $70.5$	& $82.8$	&	& $61.5$	& $61.1$	& $61.3$\\
& RecipeCrit & $\mathbf{66.3}$	& $\mathbf{74.5}$	& $\mathbf{85.4}$	&	& $\mathbf{73.7}$	& $\mathbf{74.4}$	& $\mathbf{74.1}$\\ \midrule
\multirow{4}{*}{\rotatebox{90}{\emph{Remove}}} & RecipeGPT & $91.1$	& $37.2$	& $52.9$	&	& $38.4$	& $54.6$	& $45.0$\\
& PPLM & $92.3$ & $61.3$ & $32.6$ & & $47.2$ & $53.5$ & $50.2$\\
& BART & $95.4$	& $55.7$	& $73.3$	&	& $57.6$	& $61.6$	& $59.5$\\
& RecipeCrit & $\mathbf{95.8}$	& $\mathbf{68.8}$	& $\mathbf{80.7}$	&	& $\mathbf{74.0}$	& $\mathbf{74.5}$	& $\mathbf{74.2}$ \\ %\bottomrule%We can remove it to earn space
\end{tabular}
   \caption{\label{table_crit_perf}Critiquing performance: success rate of adding/removing an ingredient,
   IoU and F1 ingredient scores, and the~Precision, Recall, and F1 of ingredients in cooking instructions.}
\end{table}

\paragraph{RQ1: Recipe Editing via Critiquing}
\label{sec_rq2}

We evaluate whether our models can edit recipes by creating new ingredient sets and corresponding recipe instructions when faced with positive and negative feedback: an ingredient that must be added or removed (substituted) from the recipe to create a new version.
For ingredient substitution, we mask the critiqued ingredient and~all steps that reference it as denoising inputs; for addition, we use the full base recipe.
For RecipeGPT and BART, we filter the predicted ingredients lists to exclude/include the target ingredient.
For PPLM, we provide the target ingredient as a bag of words to steer generation, using RecipeGPT as the base generative model.
RecipeCrit uses our iterative critiquing framework (\Cref{sec_critiquing}) to accommodate user feedback.

We show  results for constraint satisfaction (success rate), ingredient fidelity, and recipe coherence (predicted~instructions) in \Cref{table_crit_perf}.
RecipeCrit out-performs baselines across all metrics for ingredient addition and removal.
While our baselines take advantage of pre-trained language models, they cannot successfully incorporate user feedback during editing.
PPLM-guided constrained decoding is not only two orders of magnitude slower than our denoising models (3min vs. 1s per recipe), but we observe poor fidelity and frequent incoherent instructions (e.g.,~repetition).
Meanwhile, forcing ingredient lists to omit or contain specific ingredients has little impact on the generated recipe instructions---even when the desired ingredient is manually inserted into the ingredients list, RecipeGPT and BART mention using the ingredient only in 33\% and 41\% of generated~instructions.

Our model and gradient-based critiquing method leads to a stronger influence of the edited ingredients on recipe instructions.
By directly modifying the recipe latent representation that is then attended over during step generation, RecipeCrit achieves 30-50\% relative improvement in success rate for adding ingredients and 20-65\% relative improvements in coherence (F1 score between predicted ingredients and those mentioned in the instructions) for both addition and removal.
Meanwhile, baselines tend to ignore many ingredients in the ingredient list when generating new recipe directions.

\begin{table}[!t]
\small
    \centering
% \begin{threeparttable}
\begin{tabular}{@{}lcccc@{}}
\textbf{Model} & \textbf{Ser.} & \textbf{Cor.} & \textbf{Coh.} & \textbf{Rel.}\\
\toprule
RecipeGPT & $-0.04^*$ & $-0.03^*$ & $-0.01^*$ & $-0.07^*$\\
PPLM & $-0.03^*$ & $-0.05^*$ & $0.01$ & $0.00^*$\\
BART & $-0.05^*$ & $-0.07^*$ & $-0.09^*$ & $-0.07^*$\\
RecipeCrit & $\mathbf{0.12}$ & $\mathbf{0.14}$ & $\mathbf{0.10}$ & $\mathbf{0.14}$ \\ %\bottomrule%We can remove it to earn space
\end{tabular}
% \end{threeparttable}
   \caption{\label{table_hum_eval}Human evaluation of edited recipes in terms~of best-worst scaling for serendipity, correctness, coherence, and relevance. $*$ denotes a significant difference compared to RecipeCrit (posthoc Tukey test,~$p<0.01$).}%\cite{tukey}
\end{table}

\emph{Human Evaluation }
We have established that RecipeCrit creates edits that better satisfy user constraints (as expressed via critiques), 
more closely resemble the user's original preferences (base recipe), 
and make better use of the predicted ingredients (ingredient coherence).
We next perform a
qualitative
human evaluation of our edited recipes via Mechanical Turk, asking the user:
how pleasantly surprised they were (Serendipity);
whether the recipe respected their feedback (Correctness);
how easy the recipe was to follow (Coherence);
and whether the recipe resembled the original recipe (Relevance).
We uniformly sampled 800 edited recipes (400 for adding and 400 for removing) across the ingredients to critique and showed them in random order.
The annotators judged the edited recipes using best-worst scaling~\cite{louviere_flynn_marley_2015} with scores normalized to $[-1, +1]$.
\Cref{table_hum_eval} shows that our edited recipes are largely preferred on all criteria.
Our results highlight that
critiquing improves the coherence of generated recipes and their resemblance to the original~ones.

\begin{table*}[!t]
\footnotesize
\centering
\begin{tabular}{@{}p{5.4cm}@{}}
%\toprule
\multicolumn{1}{c}{\textbf{Cherry tomato confit (base recipe)}} \\
\cmidrule(lr{1em}){1-1}
clove, oil, pepper, rosemary, salt, tomato \\
\cmidrule(lr{1em}){1-1}
 1) preheat oven to 325 degrees\\
 2) spread tomatoes and garlic on a sheet. \\
 3) drizzle with oil, and sprinkle with rosemary, crushed red pepper, a large pinch of salt and several grinds of pepper.\\
 4) bake until tomatoes are wrinkled and fragrant, about 45 minutes, shaking pan.\\
 5) transfer tomato pan to a rack to cool.\\
 6) discard garlic.
\end{tabular}
\hfill
\begin{tabular}{@{}p{5.2cm}@{}}
%\toprule
\multicolumn{1}{c}{\textbf{BART}} \\
\cmidrule(lr{1em}){1-1}
\textbf{\textcolor{blue}{kale}}, cachaca, cream, ice, juice, liqueur,\\
pineapple, rum, strawberries, sugar, water\\
\textcolor{red}{\sout{clove}}, \textcolor{red}{\sout{oil}}, \textcolor{red}{\sout{pepper}}, \textcolor{red}{\sout{rosemary}}, \textcolor{red}{\sout{salt}}, \textcolor{red}{\sout{tomato}}\\
\cmidrule(lr{1em}){1-1}
 1) place ice cubes in a cocktail shaker.\\
 2) add pineapple juice, coconut liqueurs, cachacca, cream and rum ; shake well add crushed ice to a collins glass.\\
 3) add \textbf{\textcolor{blue}{kale}} and strawberries to shaker.\\
 4) strain drink into glass over crushed ice.\\
 5) garnish with strawberry and pineapple.
\end{tabular}
\hfill
\begin{tabular}{@{}p{5.0cm}@{}}
%\toprule
\multicolumn{1}{c}{\textbf{RecipeCrit (Ours)}} \\
\cmidrule(lr{1em}){1-1}
clove, \textbf{\textcolor{blue}{kale}}, oil, pepper, rosemary, salt, tomato\\
\cmidrule(lr{1em}){1-1}
 1) heat oven to 350 degrees.\\
 2) place tomatoes in a large bowl.\\
 3) drizzle with olive oil and sprinkle with rosemary, salt and pepper; coat.\\
 4) spread in a single layer on a sheet.\\
 5) roast for 40 minutes.\\
 6) remove and let cool for 10 minutes.\\
 7) toss \textbf{\textcolor{blue}{kale}} with tomatoes and garlic.
\end{tabular}
\caption{\label{tab_example}Comparison of a cherry tomato confit recipe with its edited versions to include ``\textbf{\textcolor{blue}{kale}}'' as an additional ingredient. RecipeCrit proposes tomato confit with
\textbf{\textcolor{blue}{kale}}, but BART disregards the base recipe to make a cocktail.}
\end{table*}

\emph{Case Study }
\Cref{tab_example} shows a sample of our best-performing baseline (BART) and RecipeCrit editing the ``cherry tomato confit'' recipe to include ``kale''.
While both edited recipes include kale, 
RecipeCrit stays faithful to the user's preference for ``tomato confit'' while incorporating the new feedback: it makes a slightly different tomato confit but uses kale as the ``fresh'' or salad part of the dish.
However, BART generates a \emph{cocktail} recipe instead that ignores the base recipe: 
it's a drink rather than food, sweet rather than savory, and ignores tomatoes altogether.
This aligns with the results of the human evaluation. Complementary results are shown in Table~\ref{tab_example_app} and \ref{tab_example_app_main}.

\paragraph{RQ2: Variants of Critiquing Algorithms}

Now we show the significance of the early stopping mechanism in our particular critiquing module compared to previous thresholding methods \cite{antognini2020interacting,abs-1905-12926}.
To demonstrate why, we re-run experiments from RQ1 and compare our early stopping against two baseline thresholding criteria using 1) the absolute difference (i.e., $|C(\bb{z}^*_t)_{c} - \bb{\tilde{y}}^{\mathit{ing}}_{c}| < \tau$) and 2) the L1 norm (i.e., $||C(\bb{z}^*_t) - \bb{\tilde{y}}^{\mathit{ing}}||_1 < \tau$).
We find that an L1-based stopping criterion is suboptimal due to the high dimensionality of the ingredients.
Using the absolute difference considerably improves the success rate ($+25$\% for add and $+12$\% for remove).
Finally, our early stopping further increases the success rate ($+10$\%) for both adding and removing an ingredient (see App. for exact numbers).

\section{Conclusion}
We present RecipeCrit, a denoising-based model to edit cooking recipes.
We first trained the model for recipe completion to learn semantic relationships between the ingredients and the instructions.
The novelty of this work relies on the user's ability to provide ingredient-focused feedback.
We designed an unsupervised method that substitutes the ingredients and re-writes the recipe text accordingly.
Experiments show that RecipeCrit can more effectively edit recipes compared to strong baselines, creating recipes that satisfy user constraints and are more serendipitous, correct, coherent, and relevant as measured by human judges.
For future work, we plan to extend our method to large pre-trained language models for other generative tasks and to explainable models in the context of rationalization~\cite{bastings-etal-2019-interpretable,antognini2019multi,lei-etal-2016-rationalizing,yu2021,antognini-2021-concept}.

\section{Limitations}

We demonstrated the effectiveness of our method for the English language since, to the best of our knowledge, there is no multi-lingual dataset similar to Recipe1M.
We would expect similar behavior for languages having similar morphology to English.

Regarding computational resources, the training on a single GPU takes a couple of hours, while the inference and the critiquing can run on a single-core CPU (in the range of 10 to 100 ms).

Cooking recipes are long and complex documents.
While current language models and similar ones have achieved impressive results, they still suffer from a lack of coherence for long documents.
We have shown in our experiments that RecipeCrit produced recipes whose coherence is preferred over the baselines by human annotators.
However, there is still room for improvement as language modeling approaches for recipe generation do not have an explicit guarantee of coherence (i.e.~only listed ingredients used, instructions only make use of ingredients or products mentioned before).

Similarly, as recipe instructions can consist of free-text, there is no guarantee that recipe texts will, for example, completely remove an ingredient.
In real-world usage, our system can be adapted by post-processing the recipe, including performing beam-search sampling and eliminating non-satisfactory recipes.
As a result, we continue to urge caution for users with e.g.~severe ingredient allergies who may still need to carefully review edited recipes to ensure compliance.

\bibliographystyle{acl_natbib}

\clearpage
\newpage
\appendix

\begin{algorithm*}[!t]
\caption{\label{alg_critiquing}Iterative Critiquing Gradient Update (Crit).}
\begin{algorithmic}[1]
\Function{Critique}{latent vector $\bb{z}$, critiqued ingredient $c$, trained ingredients predictor~$C$, decay coefficient~$\zeta$, patience~$P$, a maximum number of iterations $T$, desired ingredients~$\bb{\tilde{y}}^{\mathit{ing}}$}
    \State Set $\bb{z}_0 = \bb{z}^* = \bb{z}, \alpha_0 = 1, \textrm{best\_val} = \infty, \textrm{patience} = 0, t = 1$;
    \While {patience $< P$ and $t < T$}
        \State $\bb{g}_{t-1} = \nabla_{\bb{z}_{t-1}} \mathcal{L}_{\mathit{ing}}(C(\bb{z}_{t-1}), \bb{\tilde{y}}^{\mathit{ing}})$;
        \State {$\bb{z}_t = \bb{z}_{t-1} - \alpha_{t-1} \frac{\bb{g}_{t-1}}{||\bb{g}_{t-1}||_2}$ \quad and \quad $\bb{\hat{y}}^{\mathit{ing}} = C(\bb{z}_t)$}
        \If {$|\bb{\tilde{y}}^{\mathit{ing}}_{c} - \bb{\hat{y}}^{\mathit{ing}}_{c}| < \textrm{best\_val}$}
            \State {$\textrm{best\_val} = \bb{\hat{y}}^{\mathit{ing}}_{c},\bb{z}^* = \bb{z}_t,$ and $\textrm{patience} = 0$}
        \Else
            \State {$\textrm{patience} = \textrm{patience} + 1$}
        \EndIf
        
        \State {$\alpha_t = \zeta\alpha_{t-1}$\quad and \quad $t = t + 1$;}
    \EndWhile		
    \\
	\Return $\bb{z}^*$;
\EndFunction
\end{algorithmic}
\end{algorithm*}

\section{Ingredient \& Recipe Reconstruction}
\label{app_sec_rq3}

\begin{table}[!t]
    \centering
   \caption{\label{table_denoising_perf}Reconstruction performance. We report the IoU and F1 ingredient scores, and the~Precision, Recall, and F1 scores of ingredients in predicted instructions w.r.t. predicted~ones.}
% \begin{threeparttable}
\begin{tabular}{@{}l@{\hspace{1mm}}c@{\hspace{2mm}}c@{\hspace{2mm}}c@{\hspace{2mm}}c@{\hspace{2mm}}c@{\hspace{2mm}}c@{}}
& \multicolumn{2}{c}{\emph{Ingr. Fidelity}} & & \multicolumn{3}{c}{\emph{Predicted Instr.}}\\
\cmidrule(lr{1em}){2-3}\cmidrule(lr{1em}){5-7}
\textbf{Model} & \textbf{IoU} & \textbf{F1} & & \textbf{Prec.} & \textbf{Rec.} & \textbf{F1}\\
\toprule
RecipeGPT & $73.5$ & $84.7$ & & $61.2$	& $72.6$	& $66.4$\\
BART & $76.7$ & $86.4$ & & $61.5$	& $64.7$	& $63.1$\\
RecipeCrit & $\mathbf{78.6}$ & $\mathbf{88.2}$ & & $\mathbf{68.2}$	& $\mathbf{73.0}$ & $\mathbf{70.5}$ \\ %\bottomrule%We can remove it to earn space
\end{tabular}
% \end{threeparttable}
\end{table}

As baseline recipe generation models are unable to perform editing, we train all models using our denoising recipe completion task.
To evaluate their generalization performance, we ask the models to reconstruct recipes from the unseen test set, with results shown in \Cref{table_denoising_perf}.
We measure how well each model can infer the missing ingredients given the partial recipe context (IoU and F1 ingredient scores), as well as how coherent the reconstructed recipes are---the precision, recall, and F1 score of ingredients mentioned in the generated instructions compared to the predicted ingredients list.

RecipeCrit outperforms baselines in both measures.
In particular, we find a significant improvement in ingredient mention precision, indicating that RecipeCrit better constrains its generated recipe directions based on the predicted ingredients list.
Meanwhile, RecipeGPT and BART both tend to mention new ingredients in the recipe text even if they are not included in the ingredients list.
As we see in \Cref{sec_rq2}, this is problematic because such models can include ingredients in the recipe steps even if users have specified dislikes or allergies.

Such text-to-text models capture the distribution of \emph{language}~well, producing fluent-sounding text.
However, their lower scores for ingredient completion and recipe text coherence suggest that RecipeGPT and BART cannot distinguish how recipes are procedural texts with internal consistency, compared to generic text~documents.

\begin{figure}[!t]
    \centering
    \includegraphics[width=0.5\textwidth]{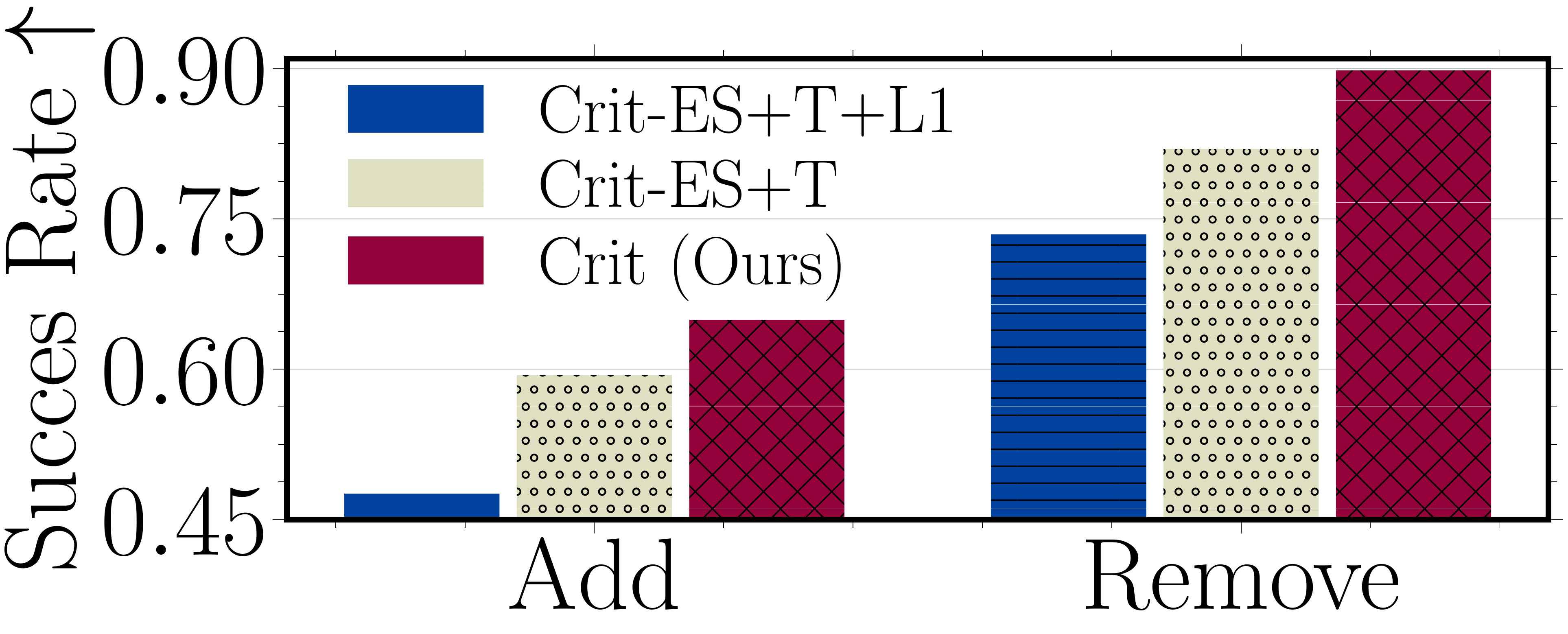}
\caption{\label{fig_crit_comp}Critiquing algorithm comparison between global or local stopping criteria with threshold or early stopping.}
\end{figure}

\section{Additional Training Details}
\label{app_training}

We use a batch~size of $32$, dropout of $0.2$, and Adam with learning rate $0.0001$. For the baselines RecipeGPT and PPLM, we reuse the official code from the authors. For BART, we employ the HuggingFace library.

\subsection{Hardware / Software}

\begin{itemize}
  \item \textbf{CPU}: 2x Intel Xeon E5-2680 v3, 2x 12 cores, 24 threads, 2.5 GHz, 30 MB cache;
  \item \textbf{RAM}: 16x16GB DDR4-2133;
  \item \textbf{GPU}: 1x Nvidia Titan X Maxwell;
  \item \textbf{OS}: Ubuntu 18.04;
  \item \textbf{Software}: Python 3.6, PyTorch 1.6.1, CUDA 10.2.
\end{itemize}

\begin{table*}[!t]
\footnotesize
\centering
\caption{\label{tab_example_app}Comparison of a cherry tomato confit recipe with its edited versions to include ``\textbf{\textcolor{blue}{kale}}'' as an additional ingredient. None of those baselines add the critique ingredient. RecipeGPT disregards the base recipe. PPLM generates a recipe in the same style than the base one, but three times longer and with a lack of coherence.}
\begin{tabular}{@{}p{7.5cm}@{}}
%\toprule
\multicolumn{1}{c}{\textbf{Cherry tomato confit (base recipe)}} \\
\cmidrule(lr{1em}){1-1}
clove, oil, pepper, rosemary, salt, tomato \\
\cmidrule(lr{1em}){1-1}
 1) preheat oven to 325 degrees\\
 2) spread tomatoes and garlic on a sheet. \\
 3) drizzle with oil, and sprinkle with rosemary, crushed red pepper, a large pinch of salt and several grinds of pepper.\\
 4) bake until tomatoes are wrinkled and fragrant, about 45 minutes, shaking pan.\\
 5) transfer tomato pan to a rack to cool.\\
 6) discard garlic.
\end{tabular}
\hfill
\begin{tabular}{@{}p{7.5cm}@{}}
%\toprule
\multicolumn{1}{c}{\textbf{RecipeGPT}} \\
\cmidrule(lr{1em}){1-1}
cream, cheese, margarine, milk, onion, pepper, salmon,\\
salt, tea\\
\textbf{\textcolor{red}{\sout{clove}}}, \textbf{\textcolor{red}{\sout{oil}}}, \textbf{\textcolor{red}{\sout{rosemary}}}, \textbf{\textcolor{red}{\sout{tomato}}}\\ 
\cmidrule(lr{1em}){1-1}
1) saute green peppers and onions in margarine.\\
2) add cream cheese and stir until melted.\\
3) add velveeta, milk and corn.\\
4) heat until melted, stirring occasionally.\\
5) add salt.\\
6) garnish with thin strips of red pepper and cilantro.\\
7) serve with crackers.\\
\end{tabular}
\hfill
\begin{tabular}{@{}p{16cm}@{}}
%\toprule
\multicolumn{1}{c}{\textbf{PPLM}} \\
\cmidrule(lr{1em}){1-1}
oil, pepper, tomato, 
\textbf{\textcolor{red}{\sout{clove}}}, \textbf{\textcolor{red}{\sout{rosemary}}}, \textbf{\textcolor{red}{\sout{tomato}}}\\ 
\cmidrule(lr{1em}){1-1}
1) preheat the oven to 400 degrees f. \\
2) place a large roasting pan in the center of the oven. \\
3) heat the olive oil in a large skillet over medium heat until hot. \\
4) add the tomatoes and cook for about 3 minutes or until the skins begin to split. \\
5) place in the oven for about 30 to 40 minutes until the skins begin to blister and shrivel. \\
6) place the tomatoes in cold water to cool, then remove the skins. \\
7) in a bowl, combine the tomatoes and their juices, salt, and pepper, tossing well to coat evenly. \\
8) remove the tomatoes, place them on a sheet tray, and bake for about 45 minutes, until the skins begin to blister and shrivel. \\
9) the tomatoes should still be juicy, and the skins should be very wrinkly. remove the tomatoes from the oven and let cool. \\
10) remove the skins from the tomatoes and discard the skins. transfer the tomatoes to a clean food processor. \\
11) blend the tomatoes with the roasted garlic, and add the peppercorns in the last 3 minutes of the cooking time, just until you can no longer process the tomatoes into a paste. place a sheet of parchment paper 12) on a clean work surface. \\
13) place a sheet of the phyllo on the work surface. \\
14) (keep the rest of the oil covered with damp paper towel. \\
15) brush the sheet of phyllo with oil. repeat with two more sheets of phyllo to form the second sheet. \\
16) you should have four layers. \\
17) if you have extra sheets left, place on a baking sheet and toast in oven with the oven door propped open until the sheets become golden brown. \\
18) remove the phyllo sheets from the oven and place on a clean work surface. cut each layered sheet along the scored lines into quarters. \\
19) place 1 heaping tablespoon of the tomato-peppercorn mixture in the center of each of four phyllo sheets. \\
20) fold the phyllo over into a triangle and place on a parchment-lined baking sheet. bake at 400 degrees f for 7 to 8 minutes, until lightly browned. \\
21) cool slightly before removing from the baking sheet, and serve warm.\\
\end{tabular}
\end{table*}

\begin{table*}[!t]
\footnotesize
\centering
\caption{\label{tab_example_app_main}Comparison of a baked parmesan fish recipe with its edited versions to include ``\textbf{\textcolor{blue}{cream}}'' as an additional ingredient. RecipeCrit incorporates the cream into a single-batter style fry, while BART only uses cream as a ``serve with sour cream'' addition to the original recipe. RecipteGPT  disregards the base recipe to make a pizza. PPLM generates a completely different and incoherent recipe, and it suffers from text degeneration.}
\begin{tabular}{@{}p{5.4cm}@{}}
%\toprule
\multicolumn{1}{c}{\textbf{Baked parmesan fish (base recipe)}} \\
\cmidrule(lr{1em}){1-1}
catfish, cheese, egg, flour, milk, paprika, pepper, salt\\
\cmidrule(lr{1em}){1-1}
1) preheat oven to 350 degrees.\\
2) rinse fillets and pat dry with paper towel.\\
3) in a pie plate , combine parmesan cheese , flour , paprika , salt and pepper.\\
4) in another pie plate , combine milk and beaten egg.\\
5) dip fillet in egg mixture and then coat with parmesan cheese mixture.\\
6) place in a well greased 9 x 13 pan.\\
7) bake uncovered for 25-30 minutes or until fish easily flakes with a fork.\\
\end{tabular}
\hfill
\begin{tabular}{@{}p{5.2cm}@{}}
%\toprule
\multicolumn{1}{c}{\textbf{BART}} \\
\cmidrule(lr{1em}){1-1}
\textbf{\textcolor{blue}{cream}}, egg, fillets, flour, garlic, milk, paprika, parsley, pepper, salt, thyme\\
\textbf{\textcolor{red}{\sout{catfish}}}, \textbf{\textcolor{red}{\sout{cheese}}}\\
\cmidrule(lr{1em}){1-1}
1) preheat oven to 350 degrees.\\
2) rinse fillets and pat dry with paper towel.\\
3) in a pie plate, combine parmesan cheese, flour, paprika, salt and in another pie plate or bowl, combine milk and beaten egg.\\
4) dip fillet in egg mixture and then coat with parmesa cheese mixture.\\
5) place in a well greased 9 x 13 pan.\\
6) bake uncovered for 25-30 minutes or until fish easily flakes with a fork.\\
7) remove from oven and let stand for 5 minutes.\\
8) serve with sour \textbf{\textcolor{blue}{cream}} and lemon wedges.\\
9) enjoy!\\
\end{tabular}
\hfill
\begin{tabular}{@{}p{5.0cm}@{}}
%\toprule
\multicolumn{1}{c}{\textbf{RecipeCrit (Ours)}} \\
\cmidrule(lr{1em}){1-1}
catfish, cheese, \textbf{\textcolor{blue}{cream}}, egg, flour, paprika, pepper, salt\\
\textbf{\textcolor{red}{\sout{milk}}}\\
\cmidrule(lr{1em}){1-1}
1) preheat oven to 350 degrees f ( 175 degrees c ).\\
2) grease a baking dish.\\
3) rinse fillets and pat dry with paper towel.\\
4) in a medium bowl , mix together the \textbf{\textcolor{blue}{cream}} cheese , sour \textbf{\textcolor{blue}{cream}} , flour , salt , paprika , and pepper.\\
5) beat in the eggs.\\
6) dip filles\\
7) pour the mixture into the baking dish.\\
8) bake in the preheated oven until the fish flakes easily with a fork , about 30 minutes.\\
\end{tabular}\\
\begin{tabular}{@{}p{7.0cm}@{}}
%\toprule
\multicolumn{1}{c}{\textbf{RecipeGPT}} \\
\cmidrule(lr{1em}){1-1}
crusts, jalapeno, mushrooms, oil, popcorn, sauce, tomato\\
\textbf{\textcolor{red}{\sout{catfish}}}, \textbf{\textcolor{red}{\sout{cheese}}}, \textbf{\textcolor{red}{\sout{egg}}}, \textbf{\textcolor{red}{\sout{flour}}}, \textbf{\textcolor{red}{\sout{milk}}}, \textbf{\textcolor{red}{\sout{paprika}}}, \textbf{\textcolor{red}{\sout{pepper}}}, \textbf{\textcolor{red}{\sout{salt}}}\\
\cmidrule(lr{1em}){1-1}
1) combine barbecue sauce and tomato sauce in bowl ; set aside.\\
2) toss mushrooms with oil ; bake in 425 degrees f standard oven 5 min.\\
3) or until softened.\\
4) place crusts on sheet pan.\\
5) spread each crust with 1/2 cup sauce mixture.\\
6) top evenly with mushrooms, jalapenos, shrimp and cheese.\\
7) bake 7 min.\\
8) or until cheese is melted and golden brown.\\
9) slice each pizza into 8 pieces.\\
10) serve with remaining barbecue sauce mixture.\\
11) serve with pickled jalapeno slices.\\
\end{tabular}\hfill
\begin{tabular}{@{}p{8.0cm}@{}}
%\toprule
\multicolumn{1}{c}{\textbf{PPLM}} \\
\cmidrule(lr{1em}){1-1}
basil, bread, butter, cod, garlic, eggplant, oil, onions, parmesan, parmigiano, parsley, pepper, salt, tomatoes\\
\textbf{\textcolor{red}{\sout{catfish}}}, \textbf{\textcolor{red}{\sout{cheese}}}, \textbf{\textcolor{red}{\sout{egg}}}, \textbf{\textcolor{red}{\sout{flour}}}, \textbf{\textcolor{red}{\sout{milk}}}, \textbf{\textcolor{red}{\sout{paprika}}}, \textbf{\textcolor{red}{\sout{pepper}}}\\ 
\cmidrule(lr{1em}){1-1}
1) preheat oven to 400 degrees. \\
2) wash the potatoes and put them in a saucepan with the \textbf{\textcolor{blue}{cream}}. \\
3) bring \textbf{\textcolor{blue}{cream}} to a boil and add the parmigiano-\textbf{\textcolor{blue}{cream}} cheese and the \textbf{\textcolor{blue}{cream}} and \textbf{\textcolor{blue}{cream}}. \\
4) let simmer, stirring often, to make a soft creamy \textbf{\textcolor{blue}{cream}} and thick \textbf{\textcolor{blue}{cream}} cheesecreamcream. \\
5) season with salt and pepper set aside. melt the cream cheese and \textbf{\textcolor{blue}{cream}} togethercream and \textbf{\textcolor{blue}{cream}} together. \\
6) beat in the buttercreamcreamcreamcreamcreamcream is verycreamcreamcreamcreamcreamcreamcreamcream.\\
7) creamcreamcreamcreamcreamcreamcreamcreamingcreamcreamcreamcreamcreamcreamcreamcream.\\\end{tabular}
\end{table*}

\end{document}